\documentclass[conference]{IEEEtran}
\pdfoutput=1
\IEEEoverridecommandlockouts
\usepackage{cite}
\usepackage{amsmath,amssymb,amsfonts}
\usepackage{algorithmic}
\usepackage{graphicx}
\usepackage{booktabs}
\usepackage{multirow}
\usepackage[caption=false]{subfig}
\usepackage{textcomp}
\usepackage{xcolor}
\usepackage[colorlinks,linkcolor=blue]{hyperref}
\def\BibTeX{{\rm B\kern-.05em{\sc i\kern-.025em b}\kern-.08em
    T\kern-.1667em\lower.7ex\hbox{E}\kern-.125emX}}
\begin{document}

\title{STNet: Spatial and Temporal feature fusion network for change detection in remote sensing images
}

\author{
\IEEEauthorblockN{\textit{Xiaowen Ma}$^{1}$,
\textit{Jiawei Yang}$^{1}$, 
\textit{Tingfeng Hong}$^{1}$,
\textit{Mengting Ma}$^{1}$,
\textit{Ziyan Zhao}$^{1}$,
\textit{Tian Feng}$^{1,2}$\IEEEauthorrefmark{1}\thanks{\IEEEauthorrefmark{1}~Corresponding author (Email: t.feng@zju.edu.cn).\newline This work was supported in part by the National Natural Science Foundation of China under Grant No. 62202421; in part by Zhejiang Provincial Key Research and Development Program under Grant No. 2021C01031; in part by Zhejiang Provincial Natural Science Foundation of China under Grant No. LTGS23F020001; in part by Ningbo Yongjiang Talent Introduction Programme under Grant No. 2021A-157-G; and in part by the Public Welfare Science and Technology Plan of Ningbo City under Grant No. 2022S125.} and
\textit{Wei Zhang}$^{1}$
}

\IEEEauthorblockA{$^{1}$Zhejiang University \quad $^{2}$Alibaba-Zhejiang University Joint Research Institute of Frontier Technologies}

}



\maketitle

\begin{abstract}
As an important task in remote sensing image analysis, remote sensing change detection (RSCD) aims to identify changes of interest in a  region from spatially co-registered multi-temporal remote sensing images, so as to monitor the local development. Existing RSCD methods usually formulate RSCD as a binary classification task, representing changes of interest by merely feature concatenation or feature subtraction and recovering the spatial details via densely connected change representations, whose performances need further improvement. In this paper, we propose STNet, a RSCD network based on spatial and temporal feature fusions. Specifically, we design a temporal feature fusion (TFF) module to combine bi-temporal features using a cross-temporal gating mechanism for emphasizing changes of interest; a spatial feature fusion module is deployed to capture fine-grained information using a cross-scale attention mechanism for recovering the spatial details of change representations. Experimental results on three benchmark datasets for RSCD demonstrate that the proposed method achieves the state-of-the-art performance. Code is available at~\href{https://github.com/xwmaxwma/rschange}{https://github.com/xwmaxwma/rschange}.
\end{abstract}


\section{Introduction}
Object-level change detection is a fundamental and crucial task in the area of remote sensing, focusing on discriminating the spectral alterations because of the changed objects that interest the users. In particular, remote sensing change detection (RSCD) processes a series of spatially co-registered images on a region but taken at different times, and aims to generate a change map of interest, whose each pixel is labelled about if the corresponding area in the region belongs to changes of interest. It reveals the local development from a natural/socioeconomic perspective and can prominently contribute to urban planning, environmental monitoring and disaster assessment~\cite{use2,use3}.

\begin{figure}[t]
	\centering
	\subfloat[]{
    \begin{minipage}[t]{0.3\linewidth}
    \includegraphics[width=1\linewidth]{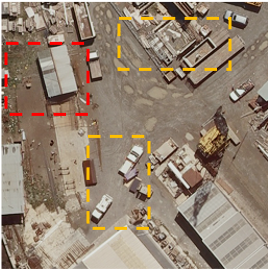}\vspace{4pt}
    \end{minipage}}
	\subfloat[]{
    \begin{minipage}[t]{0.3\linewidth}
    \includegraphics[width=1\linewidth]{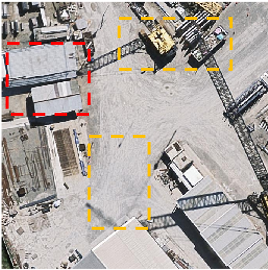}\vspace{4pt}
    \end{minipage}}
	\subfloat[]{
    \begin{minipage}[t]{0.3\linewidth}
    \includegraphics[width=1\linewidth]{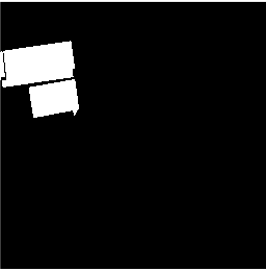}\vspace{4pt}
    \end{minipage}}
	\caption{An example of remote sensing change detection. (a) and (b) are the bi-temporal remote sensing images; (c) is the ground truth. The objects in the red dotted box are changes of interest (e.g., buildings), and those in the orange dotted box are changes of non-interest (e.g., vehicles, devices).}
	\label{fig:example}
\end{figure}

Existent RSCD methods have been developed in connection with the basic analysis units, i.e., generating difference images by common transformations, such as change vector analysis and Markov random field models, followed by thresholding or cluster analysis to obtain change representations~\cite{peng2020optical, liu2021change}. However, these methods are yet to capably detect the critical information on the changes in remote sensing images, due to the reliance on hand-crafted features. The recent rise of deep learning enables convolutional neural networks (CNN) to facilitate the evolution of RSCD methods, based on the extraction of discriminative hierarchical features. For example, IFNet~\cite{zhang2020deeply} and DTCDSCN~\cite{liu2020building} adopt a U-shaped architecture to obtain change maps by concatenating bi-temporal image features at different spatial scales, on which SNUNet~\cite{fang2021snunet} further employs dense connectivity to capture increased details of changes; SemiCDNet~\cite{peng2020semicdnet} and LGPNet~\cite{lgpnet} introduce attention mechanisms to provide more expressive change representations so as to improve the performance of change detection. However, these methods formulate RSCD as a binary classification task and rely on feature concatenation or feature subtraction to obtain change representations, which are densely connected at different spatial scales to recover spatial details, but there is still a lot of room for improvement regarding their performances. In addition, the design of dense connection in CNN leads to a complex structure requiring intensive computation and is hardly conducive to practical applications.

\begin{figure*}[t]
	\centering
	\includegraphics[width=0.9\textwidth]{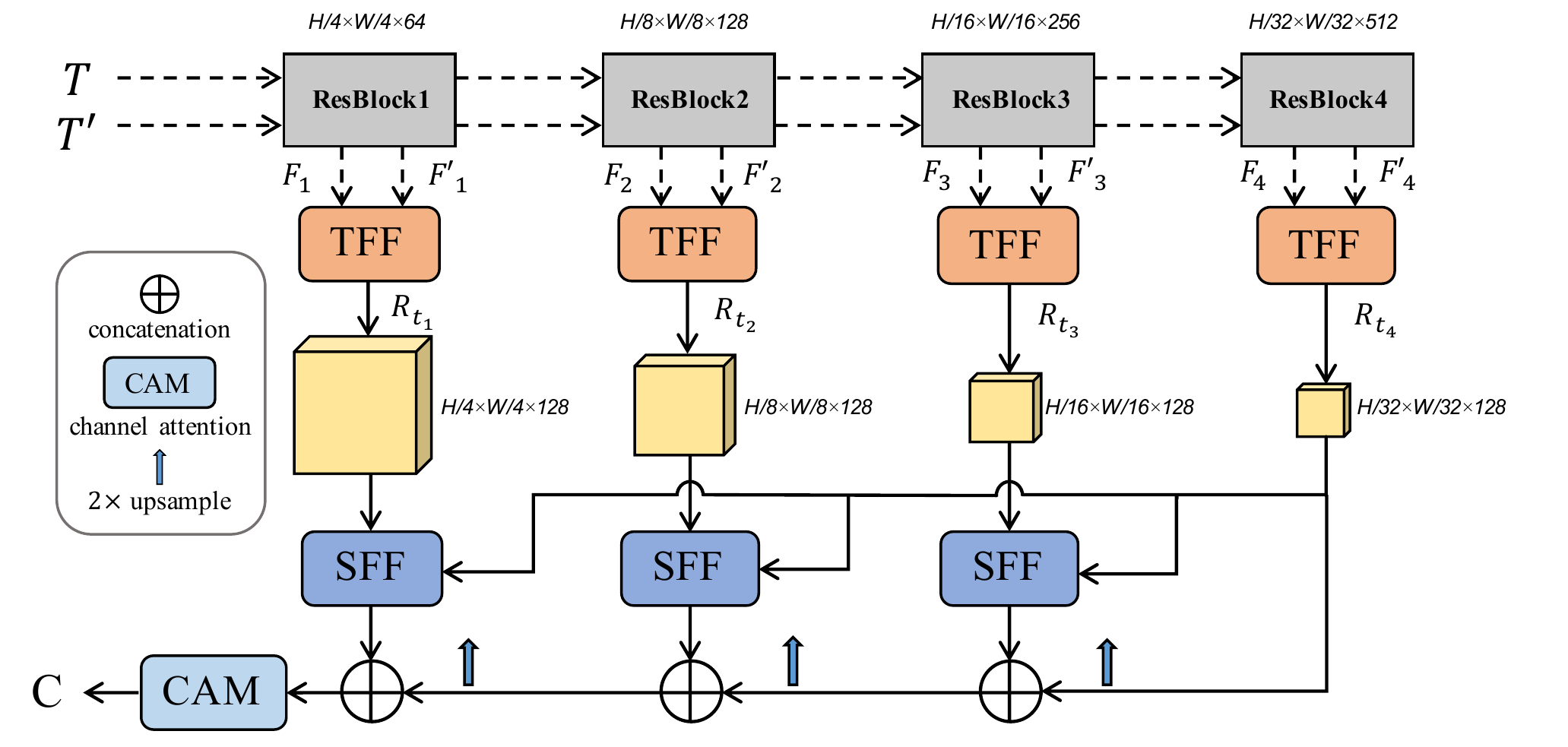}
	\caption{Architecture of the proposed STNet. From an input pair of co-registered images $\rm{T}_1$ and $\rm{T}_2$, a weight-shared ResNet-18~\cite{resnet} extracts bi-temporal features. The output of each residual block is processed by a temporal feature fusion (TFF) module into change representations, whose fine-grained spatial information is derived via cross-scale interaction based on a spatial feature fusion (SFF) module. Finally, a lightweight decoder fuses features using concatenation and a channel attention module (CAM)~\cite{senet} to obtain a change map of interest \rm{C}, representing the changes detected in $\rm{T}_1$ and $\rm{T}_2$.}
	\label{fig:whole_framework}
\end{figure*}

We argue that RSCD has the following challenges compared to general binary classification, as suggested in Fig.~\ref{fig:example}: (1) \emph{Frequent changes of non-interest}. In bi-temporal remote sensing images, distraction factors (e.g., weather, season, illumination, temporary activities) cause many spectral alterations rarely related to the natural or socioeconomic development of the region. Emphasizing changes of interest then becomes a major concern of RSCD methods; (2) \emph{High demands for spatial details}. Locating the changed objects, especially with structural information, requires efforts in RSCD due to the imbalance between unchanged and changed pixels (i.e., the number of changed pixels is usually much less than that of the unchanged). In this case, it is necessary to capture the objects' fine-grained spatial information needed for change detection.

Based on the above observations, we propose STNet, a novel spatial and temporal feature fusion network for RSCD. In particular, given the multi-scale temporal features extracted via a Siamese neural network from the input images, we design a temporal feature fusion (TFF) module and a spatial feature fusion (SFF) module to achieve informative change representations, while emphasizing changes of interest and recovering spatial details. To generate a change map of interest, we deploy a lightweight decoder that connects multi-scale change representations. 


Our contributions are summarized as follows.
\begin{itemize}
\item We present a feature fusion module that for the first time introduces a cross-temporal gating mechanism to fuse bi-temporal features for emphasizing changes of interest.

\item We present a feature fusion module that for the first time uses a cross-scale attention mechanism to capture fine-grained spatial information for recovering the spatial details of change representations.

\item We propose a deep neural network that effectively obtains multi-scale change representations, whose information can interact at different spatial scales; our method achieves the state-of-the-art performance on three benchmark datasets for RCSD in five evaluation metrics.

\item We conduct experiments on three change detection datasets. The proposed STNet significantly outperforms other state-of-the-art methods with less parameters and computational cost.
\end{itemize}

\begin{figure*}[t]
	\centering
	\includegraphics[width=0.9\textwidth]{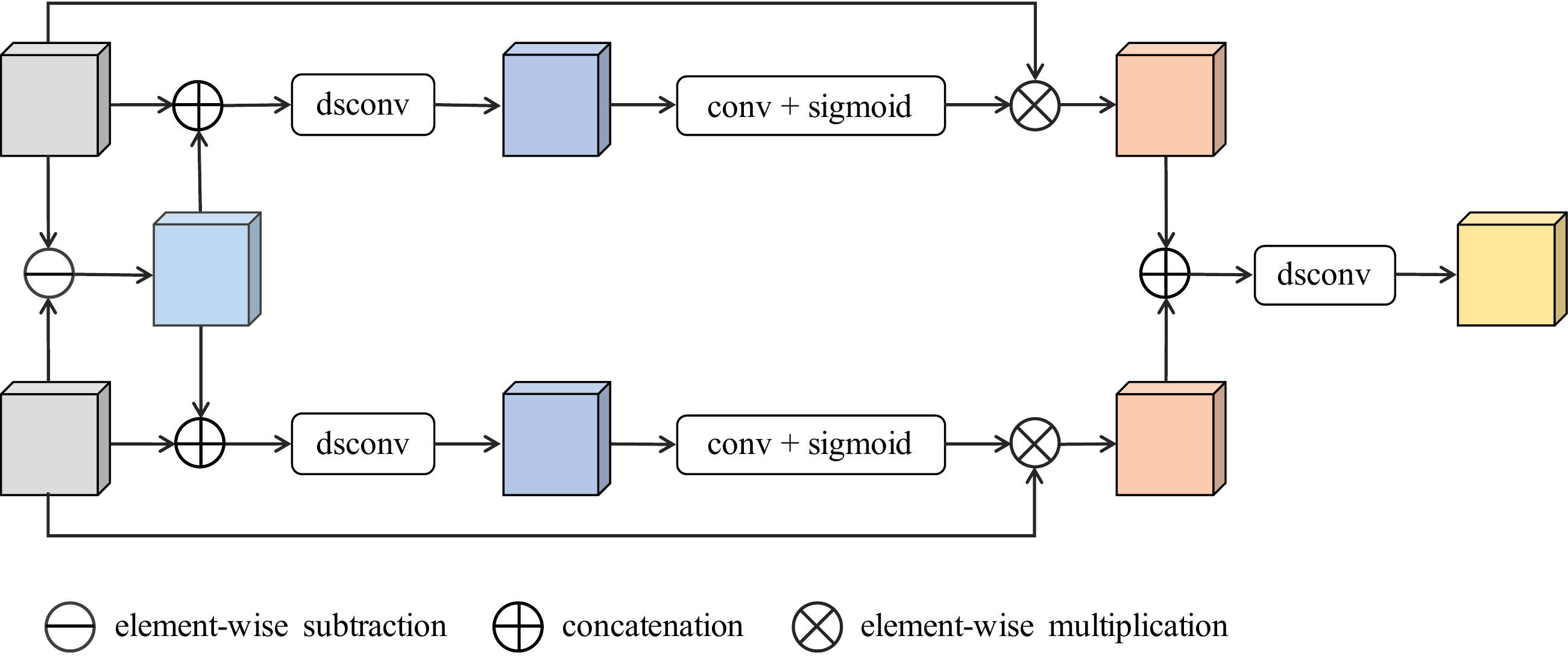}
	\caption{Structure of the TFF module, where \emph{dsconv} denotes depth-wise separable convolution for feature fusion.}
	\label{fig:tff}
\end{figure*}

\section{Method}

As illustrated in Fig.~\ref{fig:whole_framework}, the proposed STNet's workflow is divided into feature extraction, temporal feature fusion, spatial feature fusion, and decoding. From the input images, a Siamese neural network (i.e., a dual-stream ResNet-18) extracts multi-scale bi-temporal features. At each scale, a TFF module processes bi-temporal features output from the corresponding residual block to emphasize changes of interest. The features fused at the last scale are processed along with their counterpart by a SFF module at each of other scales, so as to capture the fine-grained spatial information representing the changes and recover the spatial details. Feature maps from all scales are upsampled and stitched along the channel direction, followed by feature enhancement using a channel attention module (CAM); The final upsampling on the enhanced feature map outputs a change map of interest with the same size as the input images.

\subsection{Temporal Feature Fusion}
Bi-temporal remote sensing images have complex ground objects and inexactly consistent imaging conditions (e.g., weather, season, illumination, temporary activities): the former leads to objects with the same semantic exhibiting distinct spectral characteristics, while the latter results in producing various spectral alterations. These factors enable a large number of changes of non-interest escalating the difficulty in fusing bi-temporal features for change detection. Most of the previous methods rely on feature concatenation or feature subtraction only, while rarely considering the interference of changes of non-interest. Therefore, these methods perform incompetently on adapting themselves to different RSCD objectives. 

In contrast, our temporal feature fusion (TFF) module adopts a cross-temporal gating mechanism to guide the change-specific processing (i.e., selectively enhancing changes of interest while suppressing those of non-interest), which achieves high-quality bi-temporal feature fusion. As shown in Fig.~\ref{fig:tff}, bi-temporal features $\mathcal{R}_1$ and $\mathcal{R}_2$ are firstly subtracted to obtain coarse change representations $\mathcal{R}_c$ as follows:
\begin{equation}
\mathcal{R} _c = \mathcal{R}_1 \ominus \mathcal{R}_2,
\end{equation}
where $\ominus$ denotes element-wise subtraction. $\mathcal{R} _c$ is then concatenated with $\mathcal{R}_1$ and $\mathcal{R}_2$, respectively, followed by depth-wise separable convolution to obtain $\mathcal{R}_{c1}$ and $\mathcal{R}_{c2}$ as follows:
\begin{equation}
\begin{split}
&\mathcal{R} _{c1} = \psi(\mathcal{R}_1 \oplus \mathcal{R}_c),\\
&\mathcal{R} _{c2} = \psi(\mathcal{R}_2 \oplus \mathcal{R}_c),
\end{split}
\end{equation}
where $\oplus$ denotes concatenation and $\psi$ represents depth-wise separable convolution. Note that depth-wise separable convolution replaces normal convolution here to substantially reduce the number of parameters as well as the computational cost without affecting the performance. This design enables our method to be lightweight. The weights of $\mathcal{R}_{c1}$ and $\mathcal{R}_{c2}$ are calculated as follows:
\begin{equation}
\begin{split}
&\mathcal{W} _1 = \sigma(\varphi(\mathcal{R} _{c1} )),\\
&\mathcal{W} _2 = \sigma(\varphi(\mathcal{R} _{c2} )),
\end{split}
\end{equation}
where $\varphi$ denotes $1\times1$ convolution and $\sigma$ represents a sigmoid function. Finally, the refined change representations $\mathcal{R}_t$ are calculated as follows:
\begin{equation}
\mathcal{R} _t = \psi((\mathcal{W} _1 \otimes \mathcal{R}_1) \oplus (\mathcal{W} _2 \otimes \mathcal{R}_2)),
\end{equation}
where $\otimes$ denotes element-wise multiplication.

\subsection{Spatial Feature Fusion}
The area of changed objects is generally much smaller compared to the rest in RSCD, and it is thus critical to recover the spatial details of change representations. In multi-scale change representations, the high level ones contain rich semantic information but low-quality boundaries, whereas the low level ones hold spatial details but irrelevant background information. Hence, a common strategy so far is to densely connect the change representations at different scales for semantic information and spatial details, where boundaries and background information can interfere, causing degraded performances for change detection. 

To address the above-mentioned issue, our spatial feature fusion (SFF) module relies on a cross-scale attention mechanism that exploits high-level change representations to guide low-level representations for context modeling, thus enhancing the semantic information in the latter and capturing higher-quality spatial details of change representations.

\begin{figure*}[t]
	\centering
	\includegraphics[width=0.9\textwidth]{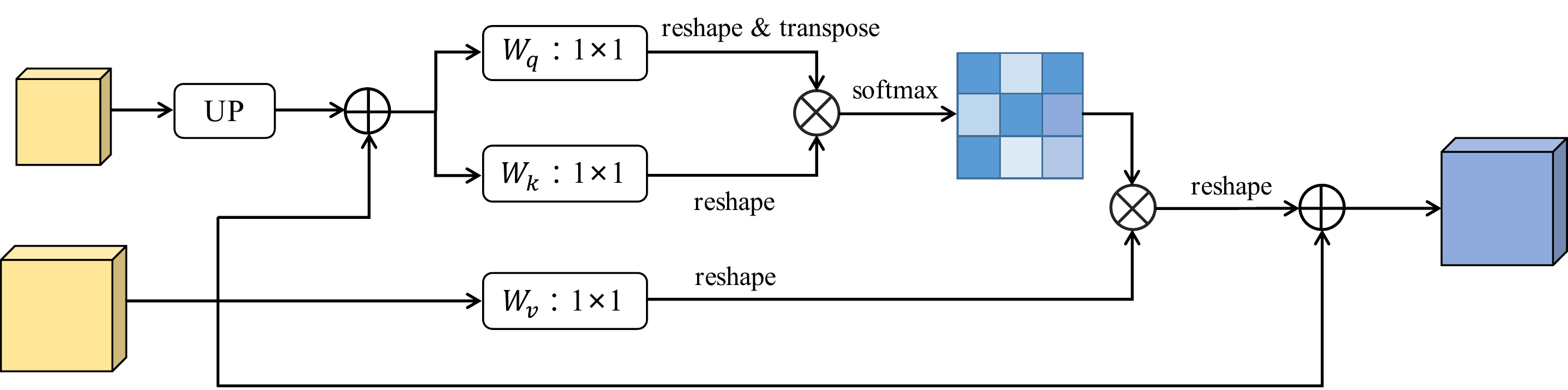}
	\caption{Structure of the SFF module, where \emph{UP} denotes upsampling via bi-linear interpolation.}
	\label{fig:sff}
\end{figure*}

\setlength{\tabcolsep}{4pt}
\begin{table*}[t]
 \begin{center}
  \caption{
      Change detection results on the three datasets. Highest scores are marked in bold. All scores are reported in percentage.
  }
  \label{table:campare}
  \begin{tabular}{l|ccccc|ccccc|ccccc}
   \toprule
   \multirow{2}{*}{Method} &\multicolumn{5}{c|}{WHU} &\multicolumn{5}{c|}{LEVIR-CD} &\multicolumn{5}{c}{CLCD}\\
     &F1 &Pre. &Rec. &IoU &OA  &F1 &Pre. &Rec. &IoU &OA &F1 &Pre. &Rec. &IoU &OA \\
   \midrule
   FC-EF~\cite{fc-siam}  &72.01 &77.69 &67.10 &56.26 &92.07 &83.4 &86.91 &80.17 &71.53 &98.39 &48.64 &73.34 &36.29 &32.14 &94.30\\
   FC-Siam-Di~\cite{fc-siam}  &58.81 &47.33 &77.66 &41.66 &95.63 &86.31 &89.53 &83.31 &75.92 &98.67 &44.10 &72.97 &31.60 &28.29 &94.04\\
   FC-Siam-Conc~\cite{fc-siam} &66.63 &60.88 &73.58 &49.95 &97.04 &83.69 &91.99 &76.77&71.96&98.49&54.48&68.21&45.22&37.35&94.35\\
   IFNet~\cite{zhang2020deeply} &83.40 &\bf96.91 &73.19 &71.52 &98.83 &88.13 &\bf94.02 &82.93&78.77&98.87&48.65&49.96&47.41&32.14&92.55\\
   DTCDSCN~\cite{liu2020building}&71.95 &63.92 &82.30 &56.19 &97.42 &87.67 &88.53&86.83&78.05&98.77&60.13&62.98&57.53&42.99&94.32 \\
   BIT~\cite{bit}&83.98 &86.64 &81.48 &72.39 &98.75 &89.31 &89.24 &\bf89.37&80.68&98.92&57.13&64.39&51.34&39.99&94.27\\
   LGPNet~\cite{lgpnet}&79.75 &89.68 &71.81 &66.33 &98.33 &89.37 &93.07 &85.95&80.78&99.00&55.42&62.98&49.49&38.33&94.10 \\
   SNUNet~\cite{fang2021snunet}&83.50 &85.60 &81.49 &71.67 &98.71 &88.16 &89.18 &87.17&78.83&98.82&60.54&65.63&56.19&43.41&94.55\\
   DMATNet~\cite{dmatnet}&85.07 &89.46 &82.24 &74.98 &95.83 &89.97 &90.78 &89.17&81.83&98.06&66.56&72.74&61.34&49.87&95.41\\
   ChangeStar(FarSeg)~\cite{changestar}&87.01 &88.78 &85.31 &77.00 &98.70 &89.30 &89.88 &88.72&80.66&98.90&60.75&62.23&59.34&43.63&94.30\\
   ChangeFormer~\cite{changeformer}&81.82 &87.25 &77.03 &69.24 &94.80 &90.40 &92.05&88.80&\bf82.48&99.04&58.44&65.00&53.07&41.28&94.38\\
   \midrule
   \textbf{STNet} (Ours)&\bf87.46 &87.84 &\bf87.08 &\bf77.72 &\bf98.85 &\bf90.52 &92.06 &89.03 &82.09 &\bf99.36&\bf73.90&\bf74.42&\bf73.38&\bf58.61&\bf96.12\\
   \bottomrule
  \end{tabular}
 \end{center}
\end{table*}
\setlength{\tabcolsep}{2pt}
To illustrate the cross-scale attention mechanism adopted in our STNet, we describe the general paradigm of attention mechanism as follows: Given the query $\emph{Q} = [\emph{Q} _{1},\emph{Q} _{2},\dots \emph{Q}_ {N}] \in \mathbb{R} ^{N\times C}$ ,where $N$ and $C$ denote the numbers of pixels and channels in $\emph{Q}$. Likewise, we have the key $\emph{K}\in \mathbb{R} ^{N\times C}$ and the value $\emph{V}\in \mathbb{R} ^{N\times C}$. Each output element $\emph{Z}_i$ is then computed as the following weighted sum of input elements:
\begin{equation}
    \emph{Z}_{i}=\sum_{j=1}^n\alpha_{ij}(\emph{V}_{j}\textbf{W}^V),
\end{equation}
where the corresponding weight coefficient $\alpha_{ij}$ is computed using the softmax faction as follows:
\begin{equation}
   \alpha_{ij}=\frac{exp(e_{ij})}{\sum\nolimits_{k=1}^nexp(e_{ik})},
\end{equation}
where $e_{ij}$ denotes a scaled dot-product attention as follows:
\begin{equation}
   e_{ij}=\frac{(\emph{Q}_{i}\textbf{W}^Q)(\emph{K}_{j}\textbf{W}^K)^T}{\sqrt{C}}.
\end{equation}
The above projections $\textbf{W}^Q$, $\textbf{W}^K$, $\textbf{W}^V$ are parameter matrices. For the self-attention mechanism, $\emph{Q}$, $\emph{K}$, $\emph{V}$ are all set to the representations at a specific layer; but for the cross-scale attention mechanism proposed in this paper, as shown in Fig.~\ref{fig:sff}, $\emph{Q}$ and $\emph{K}$ are the concatenation of the upsampled high-level change representations and low-level representations, whereas $\emph{V}$ is low-level representations. This operation enables high-level representations to provide richer and more accurate contextual information, guiding low-level representations to calculate pixel-wise relationships and enhancing their capability to represent semantic information. 

\subsection{Loss Function}
Considering the significantly unbalanced distribution of changed and unchanged pixels in RSCD, we employ a hybrid loss function combining a focal loss and a dice loss, as follows,
\begin{equation}
\mathcal{L}  =  \mathcal{L} _{focal} + \mathcal{L}_{dice}.
\end{equation}
And the focal loss is formulated as,
\begin{equation}
\mathcal{L}_{\text{focal}} = -\alpha(1-\hat{p})^{\gamma}log(\hat{p}),
\end{equation}
\begin{equation}
\hat{p}=\left \{\begin{array}{ll}p,&\text{if\quad} y=1 \\
                            1-p,&\text{otherwise} \\
                \end{array}
        \right.
\end{equation}
where $\alpha$ and $\gamma$ are two hyperparameters controlling the weights of positive and negative samples and the attention of the model to difficult samples, respectively. In the experiments, we set $\alpha$ to 0.2 and $\gamma$ to 2; $p$ is the probability and $y$ is the binary label (0 or 1) corresponding to unchanged and changed pixels.

The dice loss is formulated as,
\begin{equation}
\mathcal{L}_{dice}  =1 - \frac{2 \cdot E \cdot softmax({E}')}{E+softmax({E}')},
\end{equation}
\begin{equation}
{E}' = \left\{{e'_{k}}, \ k=1,2,\dots,H\times W\right\},
\end{equation}
where $E$ denotes the ground truth, ${E}'\in \mathbb{R} ^{H\times W \times 2}$ represents the change map and $e'_{k}\in \mathbb{R} ^2$ represents a point in ${E}'$.

\begin{figure*}[t]
\centering
\captionsetup[subfloat]{labelsep=none,format=plain,labelformat=empty}
\subfloat[$T_1$]{
\begin{minipage}[t]{0.1\linewidth}
\includegraphics[width=1\linewidth]{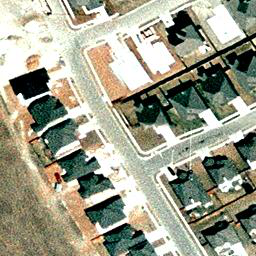}\vspace{2pt}
\includegraphics[width=1\linewidth]{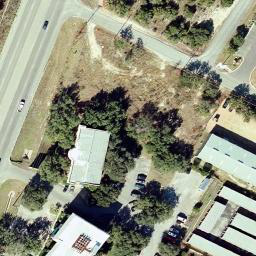}\vspace{2pt}
\includegraphics[width=1\linewidth]{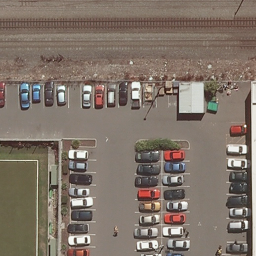}\vspace{2pt}
\includegraphics[width=1\linewidth]{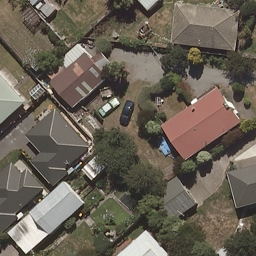}\vspace{2pt}
\end{minipage}}
\subfloat[$T_2$]{
\begin{minipage}[t]{0.1\linewidth}
\includegraphics[width=1\linewidth]{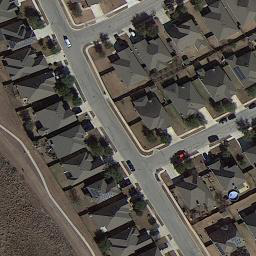}\vspace{2pt}
\includegraphics[width=1\linewidth]{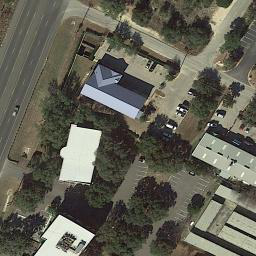}\vspace{2pt}
\includegraphics[width=1\linewidth]{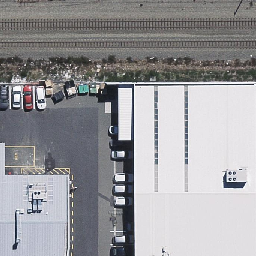}\vspace{2pt}
\includegraphics[width=1\linewidth]{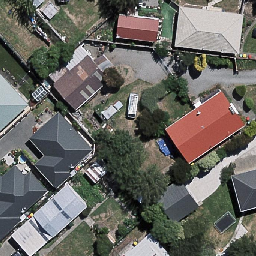}\vspace{2pt}
\end{minipage}}
\subfloat[GT]{
\begin{minipage}[t]{0.1\linewidth}
\includegraphics[width=1\linewidth]{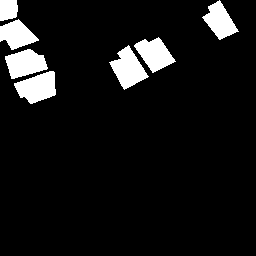}\vspace{2pt}
\includegraphics[width=1\linewidth]{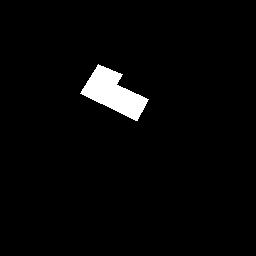}\vspace{2pt}
\includegraphics[width=1\linewidth]{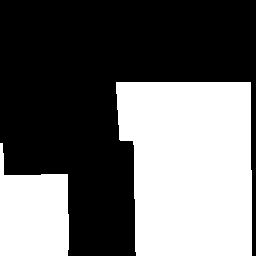}\vspace{2pt}
\includegraphics[width=1\linewidth]{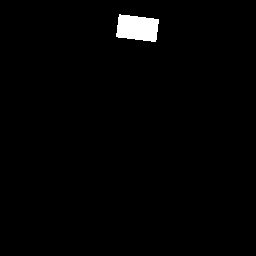}\vspace{2pt}
\end{minipage}}
\subfloat[FC-Siam-Di]{
\begin{minipage}[t]{0.1\linewidth}
\includegraphics[width=1\linewidth]{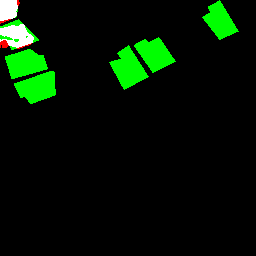}\vspace{2pt}
\includegraphics[width=1\linewidth]{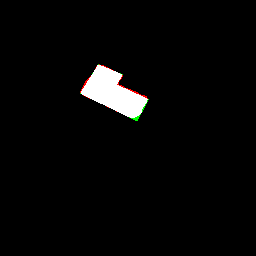}\vspace{2pt}
\includegraphics[width=1\linewidth]{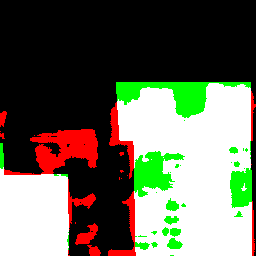}\vspace{2pt}
\includegraphics[width=1\linewidth]{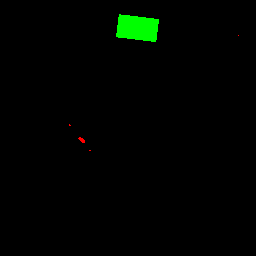}\vspace{2pt}
\end{minipage}}
\subfloat[FC-Siam-Conc]{
\begin{minipage}[t]{0.1\linewidth}
\includegraphics[width=1\linewidth]{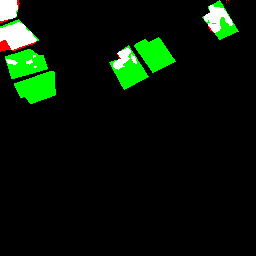}\vspace{2pt}
\includegraphics[width=1\linewidth]{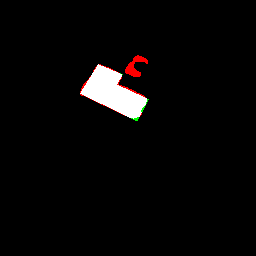}\vspace{2pt}
\includegraphics[width=1\linewidth]{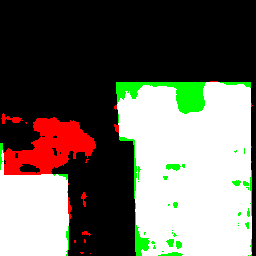}\vspace{2pt}
\includegraphics[width=1\linewidth]{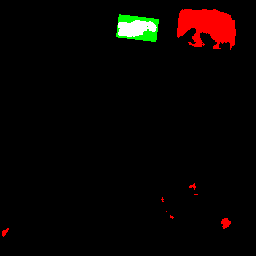}\vspace{2pt}
\end{minipage}}
\subfloat[IFNet]{
\begin{minipage}[t]{0.1\linewidth}
\includegraphics[width=1\linewidth]{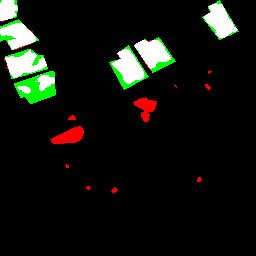}\vspace{2pt}
\includegraphics[width=1\linewidth]{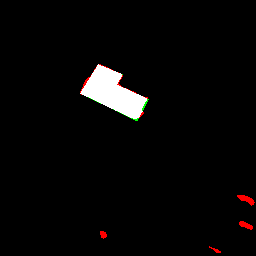}\vspace{2pt}
\includegraphics[width=1\linewidth]{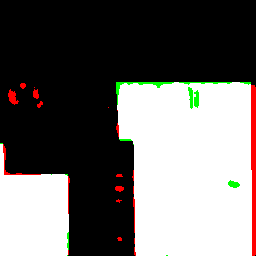}\vspace{2pt}
\includegraphics[width=1\linewidth]{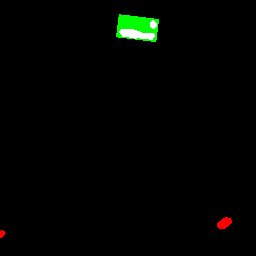}\vspace{2pt}
\end{minipage}}
\subfloat[LGPNet]{
\begin{minipage}[t]{0.1\linewidth}
\includegraphics[width=1\linewidth]{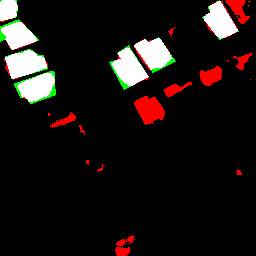}\vspace{2pt}
\includegraphics[width=1\linewidth]{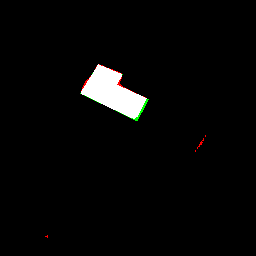}\vspace{2pt}
\includegraphics[width=1\linewidth]{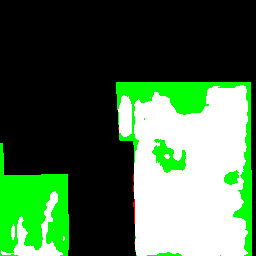}\vspace{2pt}
\includegraphics[width=1\linewidth]{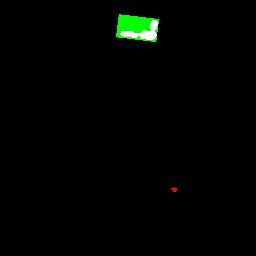}\vspace{2pt}
\end{minipage}}
\subfloat[SNUNet]{
\begin{minipage}[t]{0.1\linewidth}
\includegraphics[width=1\linewidth]{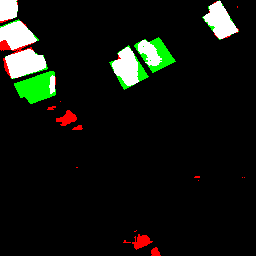}\vspace{2pt}
\includegraphics[width=1\linewidth]{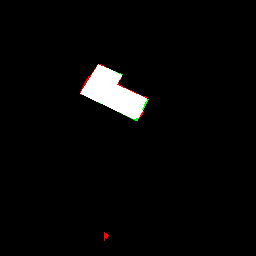}\vspace{2pt}
\includegraphics[width=1\linewidth]{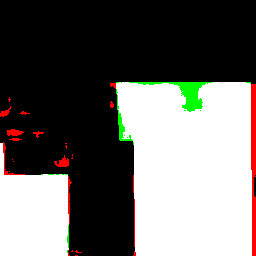}\vspace{2pt}
\includegraphics[width=1\linewidth]{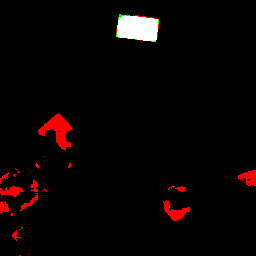}\vspace{2pt}
\end{minipage}}
\subfloat[STNet]{
\begin{minipage}[t]{0.1\linewidth}
\includegraphics[width=1\linewidth]{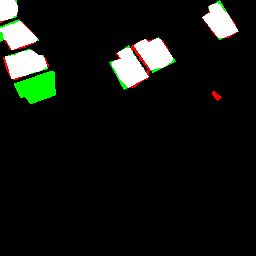}\vspace{2pt}
\includegraphics[width=1\linewidth]{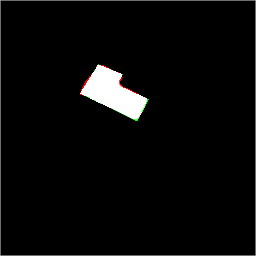}\vspace{2pt}
\includegraphics[width=1\linewidth]{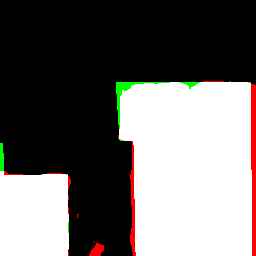}\vspace{2pt}
\includegraphics[width=1\linewidth]{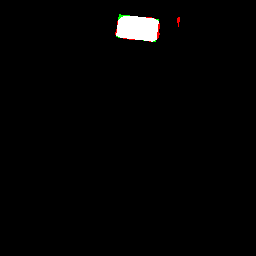}\vspace{2pt}
\end{minipage}}
\caption{Example results output from RSCD methods on test sets from WHU and LEVIR-CD. Pixels are colored differently for better visualization (i.e., white for true positive, black for true negative, red for false positive, and green for false negative).}
\label{fig:res}
\end{figure*}

\setlength{\tabcolsep}{6pt}
\begin{table}[t]
	\begin{center}
		\caption{
		      Change detection results of STNet and its variants on the WHU dataset. Highest scores are marked in bold. All scores are reported in percentage.
		}
		\label{table:abalation}
		\begin{tabular}{l|ccccc}
			\toprule
			Model &  F1&Pre.&  Rec. & IoU &OA\\
			\midrule
			 Base & 79.65& 78.14& 81.22 &66.58 &97.19 \\
			 Base+TFF & 84.20 &84.08 &84.33 &72.72 &98.55 \\
			 Base+SFF  &85.07 &87.74 &82.55 &74.01 &98.67 \\
			 \midrule
			 \textbf{STNet} (Ours) &\bf87.46 &\bf87.84 &\bf87.08 &\bf77.72 &\bf98.85 \\
			\bottomrule
		\end{tabular}
	\end{center}
\end{table}

\section{EXPERIMENTAL RESULTS}
\subsection{Datasets and Evaluation Metrics}
We evaluate the proposed STNet on three RSCD benchmark datasets in five common metrics: F1-score (F1), recall (Rec), precision (Pre), intersection over union (IoU), and overall accuracy (OA).

The WHU building dataset~\cite{whu} contains a pair of bi-temporal aerial images of $32507\times15354$ size and with 0.075~m spatial resolution. Following the previous work~\cite{bit}, we crop the images into patches of $256\times256$ size and randomly divide them into
a training set (6096 images), a validation set (762 images), and a testing set (762 images).

The LEVIR-CD~\cite{levir} is a large-scale remote sensing building change detection dataset. It contains 637 pairs of high-resolution bi-temporal remote sensing images of $1024\times 1024$ size and with 0.5~m spatial resolution. Following the dataset's default setting, we crop the images into non-overlapping patches of $256\times256$ size and randomly divide them into
a training set (7120 images), a validation set (1024 images), and a testing set (2048 images).

The CropLand Change Detection (CLCD) dataset~\cite{clcd} contains 600 pairs of remote sensing images of $512\times512$ size and with spatial resolution from 0.5~m to 2~m. We randomly divide them into a training set (360 images), a validation set (120 images), and a testing set (120 images).

\subsection{Implementation Details}
We implement the proposed STNet using Python and Pytorch on a workstation with two NVIDIA GTX A5000 graphics cards (48~GB GPU memory in total). The base learning rate is set to 1e-4 and the Adam optimizer is used with a weight decay of 1e-5. During the training period, we adopt a multi-step learning rate decay strategy with the gamma of 0.9 to update the learning rate. In the experiments, the batch size is set to 4 and we conduct data augmentation via flipping and rotating the images for training.

\setlength{\tabcolsep}{12pt}
\begin{table}[t]
	\begin{center}
		\caption{
		Comparison with the state-the-art methods on space complexity and computational cost. Input images were resized to $256\times256\times3$ for computational cost calculation.
		}
		\label{table:efficiency}
		\begin{tabular}{l|cc}
			\toprule
			Model & Params (M) &Flops (G)\\
			\midrule
			 IFNet~\cite{zhang2020deeply} & 50.44 &82.26\\
			 DTCDSCN~\cite{liu2020building} &41.07 &14.42\\
			 LGPNet~\cite{lgpnet} &70.99 &125.79 \\
			 DMATNet~\cite{dmatnet} &13.27 &- \\
			 SNUNet~\cite{fang2021snunet}&\bf12.03 &54.88 \\
			 \midrule
			 \textbf{STNet} (Ours) &14.6 &\bf9.61 \\
			\bottomrule
		\end{tabular}
	\end{center}
\end{table}

\subsection{Comparisons and Analysis}
Our experiments start with an ablation study for our STNet on the WHU dataset. Specifically, we select the variants of STNet (i.e., without TFF and SFF modules) as the base model. As shown in Table~\ref{table:abalation}, the base model achieve 79.65\% F1, while the performance is significantly improved after introducing TFF module or SFF module, proving their validity. In particular, our STNet reach the highest 87.46\% F1, showing that combining both temporal and spatial feature fusions can notably improve change detection performance.

In addition, we compare our STNet with the current state-of-the-art methods on three benchmark datasets. As shown in Table~\ref{table:campare}, our STNet achieve F1 of 87.46\%, 90.52\%, and 73.90\%, respectively, on the testing set from WHU, LEVIR-CD, and CLCD. Its improvement is significant compared to the other methods on WHU and CLCD, and is about 0.55\% higher on LEVIR-CD than a recent method DMATNet~\cite{dmatnet}. These results demonstrate our STNet's superiority on change detection performance.

Moreover, we adopt the number of parameters measured in million (refered to as \emph{Params (M)}) and the floating-point operations per second measured in giga (refered to as \emph{Flops (G)}) to measure space complexity and computational cost for our STNet and some methods for comparison. As shown in Table~\ref{table:efficiency}, the proposed method require less parameters as well as the lowest computational cost while outperforming the other methods for change detection.

We visualize example change detection results, as shown in Fig.~\ref{fig:res}. It can be observed that our STNet extract better high-level representations on the changes of interest and also recover high-quality boundaries of the changed object, because of the effectiveness of TFF and SFF modules.

\section{Conclusion}
In this paper, we analyze two outstanding challenges of remote sensing change detection and hence propose a spatial and temporal feature fusion network named STNet. Based on the cross-temporal gating mechanism and the cross-scale attention mechanism, our STNet can effectively reduce semantic ambiguity and missing spatial details of the change objects by extracting high-quality change representations and capturing their fine-grained information. The proposed method achieves the state-of-the-art performance on three benchmark datasets in the experiments, demonstrating its effectiveness for remote sensing change detection.



\bibliographystyle{IEEEtran}
\bibliography{stnet}

\end{document}